**Title:** Advancing COVID-19 Diagnosis with Privacy-Preserving Collaboration in Artificial Intelligence

**One sentence summary:** An efficient and effective privacy-preserving AI framework is proposed for CT-based COVID-19 diagnosis, based on 9,573 CT scans of 3,336 patients, from 23 hospitals in China and the UK.

**Author list**: Xiang Bai[1,2*#], Hanchen Wang[3#], Liya Ma[1#], Yongchao Xu[1#], Jiefeng Gan[2#], Ziwei Fan[2], Fan Yang[4], Ke Ma[2], Jiehua Yang[2], Song Bai[2], Chang Shu[2], Xinyu Zou[2], Renhao Huang[2], Changzheng Zhang[5], Xiaowu Liu[5], Dandan Tu[5], Chuou Xu[1], Wenqing Zhang[1], Xi Wang[6], Anguo Chen[7], Yu Zeng[8], Dehua Yang[9], Ming-Wei Wang[9], Nagaraj Holalkere[10], Neil J. Halin[10], Ihab R. Kamel[11], Jia Wu[12], Xuehua Peng[13], Xiang Wang[14], Jianbo Shao[13], Pattanasak Mongkolwat[15], Jianjun Zhang[16,17], Weiyang Liu[3], Michael Roberts[18,19], Zhongzhao Teng[20], Lucian Beer[20], Lorena Escudero Sanchez[20], Evis Sala[20], Daniel Rubin[21], Adrian Weller[3,22], Joan Lasenby[3], Chuangsheng Zheng[4*], Jianming Wang[23*], Zhen Li[1*], Carola-Bibiane Schönlieb[18,22*], Tian Xia[2*]

**Affiliations:** [1]Department of Radiology, Tongji Hospital and Medical College, Huazhong University of Science and Technology, Wuhan, China. [2]School of Artificial Intelligence and Automation, Huazhong University of Science and Technology, Wuhan, China. [3]Department of Engineering, University of Cambridge, Cambridge, UK. [4]Department of Radiology, Union Hospital of Tongji Medical College, Huazhong University of Science and Technology, Wuhan, China. [5]HUST-HW Joint Innovation Lab, Wuhan, China. [6]CalmCar Inc, Suzhou, China. [7]Wuhan Blood Centre, Wuhan, China. [8]MSA Capital, Beijing, China. [9]The National Centre for Drug Screening, Shanghai Institute of Materia Medica, Chinese Academy of Sciences, Shanghai, China. [10]CardioVascular and Interventional Radiology, Radiology for Quality and Operations, The Cardiovascular Centre at Tufts Medical Centre, Radiology, Tufts University School of Medicine, Medford, USA. [11]Russell H Morgan Department of Radiology & Radiologic Science, Johns Hopkins Hospital & Medicine Institute, Baltimore, USA. [12]Department of Radiation Oncology, School of Medicine, Stanford University, Palo Alto, USA. [13]Department of Radiology, Wuhan Central Hospital, Wuhan, China. [14]Department of Radiology, Wuhan Children's Hospital, Wuhan, China. [15]Faculty of Information and Communication Technology, Mahidol University, Thailand. [16]Thoracic/Head and Neck Medical Oncology, [17]Translational Molecular Pathology, University of Texas MD Anderson Cancer Centre, Houston, USA. [18]Department of Applied Mathematics and Theoretical Physics, University of Cambridge, Cambridge, UK. [19]Oncology R&D at AstraZeneca, Cambridge, UK. [20]Department of Radiology, University of Cambridge, Cambridge, UK. [21]Department of Biomedical Data Science, Radiology and Medicine, Stanford University, Palo Alto, USA. [22]Alan Turing Institute, London, UK. [23]Department of Hepatobiliary Pancreatic Surgery, Affiliated Tianyou Hospital, Wuhan University of Science and Technology, Wuhan, China.

# These authors contributed equally to this work. * Correspondence can be addressed to X.B. (xbai@hust.edu.cn), C.Z. (cszheng@163.com), J.W. (jmwang@163.com), Z.L. (zhenli@hust.edu.cn), C.B. (cbs31@cam.ac.uk), and T.X. (tianxia@hust.edu.cn).

**Abstract**

Artificial intelligence (AI) provides a promising substitution for streamlining COVID-19 diagnoses. However, concerns surrounding security and trustworthiness impede the collection of large-scale representative medical data, posing a considerable challenge for training a well-generalised model in clinical practices. To address this, we launch the Unified CT-COVID AI Diagnostic Initiative (UCADI), where the AI model can be distributedly trained and independently executed at each host institution under a federated learning framework (FL) without data sharing. Here we show that our FL model outperformed all the local models by a large yield (test sensitivity /specificity in China: 0.973/0.951, in the UK: 0.730/0.942), achieving comparable performance with a panel of professional radiologists. We further evaluated the model on the hold-out (collected from another two hospitals leaving out the FL) and heterogeneous (acquired with contrast materials) data, provided visual explanations for decisions made by the model, and analysed the trade-offs between the model performance and the communication costs in the federated training process. Our study is based on 9,573 chest computed tomography scans (CTs) from 3,336 patients collected from 23 hospitals located in China and the UK. Collectively, our work advanced the prospects of utilising federated learning for privacy-preserving AI in digital health.

## MAIN TEXT

### Introduction

As the gold standard for identifying COVID-19 carriers, reverse transcription-polymerase chain reaction (RT-PCR) is the primary diagnostic modality to detect viral nucleotide in specimens from cases with suspected infection. However, due to the various disease courses in different patients, the detection sensitivity hovers at around only 0.60 – 0.71[1–4], which results in a considerable number of false negatives. As such, clinicians and researchers have made tremendous efforts searching for alternatives[5–7] and complementary modalities[2,8–11] to improve the testing scalability and accuracy for COVID-19 and beyond.

It has been reported that coronavirus carriers present certain radiological features in chest CTs, including ground-glass opacity, interlobular septal thickening, and consolidation, which can be exploited to identify COVID-19 cases. Chest CTs have thus been utilised to diagnose COVID-19 in some countries and regions with reported sensitivity ranging from 0.56 to 0.98[12–15]. However, these radiological features are not explicitly tied to COVID-19, and the accuracy of CT-based diagnostic tools heavily depends on the radiologists' own knowledge and experience. A recent study[16] has further investigated the substantial discrepancies in differentiating COVID-19 from other viral pneumonia by different radiologists. Such inconsistency is undesirable for any clinical decision system. Therefore, there is an urgent demand to develop an accurate and automatic method to help address the clinical deficiency in current CT-based approaches.

Successful development of an automated method relies on a sufficient amount of data accompanied by precise annotations. We identified three challenges, specifically data-related, for developing a robust and generalised AI model for CT-based COVID-19 identifications: **(i) Incompleteness**. High-quality CTs that were used for training was only a small subset of the entire cohort; therefore, they are unlikely to cover the complete set of useful radiological features for identification. **(ii) Isolation**. CTs acquired across multiple centres were difficult to transfer for training due to security and privacy concerns, while a locally trained model may not be generalised to or improved by the data collected from other sites. **(iii) Heterogeneity**. Due to the different acquisition protocols (e.g., contrast agents and reconstruction kernels), CTs collected from a single hospital are still not yet well standardised; therefore, it is challenging to train a precise model based on a simple combination of the data[17].

Furthermore, it remains an open question whether the COVID-19 patients from diverse geographies and varying demographics show similar or distinct patterns. All these challenges will impede the development of a well-generalised AI model, and thus, of a global intelligent clinical solution. It is worth noting that these challenges are generally encountered by all the possible trails in applying AI models in clinical practices, not necessarily COVID-19 related.

To tackle these problems, we launched the Unified CT-COVID AI Diagnostic Initiative (UCADI, in Fig. 1 and 2). It was developed based on the concept of federated learning[18,19], which enables machine learning engineers and clinical data scientists to collaborate seamlessly, yet without sharing the patient data. Thus, in UCADI, every participating institution can benefit from, and contribute to, the continuously evolving AI model, helping deliver even more precise diagnoses for COVID-19 and *beyond*.

## Results

- **Developing a local accurate AI diagnostic model**

Training an accurate AI model requires comprehensive data collection. Therefore, we first gathered, screened, and anonymized the chest CTs at each UCADI participating institute (5 hospitals in China and 18 hospitals in the UK), comprising a total of 9,573 CTs of 3,336 cases. We summarised the demographics and diagnoses of the cohort in the Supplementary Table 1 and 2.

Developing an accurate diagnostic model requires a sufficient amount of high-quality data. Consequently, we identified the three branches of Wuhan Tongji Hospital Group (Main Campus, Optical Valley and Sino-French) and the National COVID-19 Chest Imaging Database (NCCID)[20] as individual UCADI participants. Each site contains adequate high-quality CTs for the development of the 3D CNN model. We used 80% of the data for training and validation (trainval) and the rest 20% for testing. Additionally, we utilize the CTs collected from Tianyou hospital and Wuhan Union hospital as hold-out test sets. We consistently use the same partition in both the local and federated training processes for a fair comparison.

NCCID is an initiative established by NHSX, providing massive CT and CXR modalities of COVID-19 and non-COVID-19 patients from over 18 partnership hospitals in the UK. Since each hospital's data quantity and categorial distribution are quite uneven, we pooled all the CTs and identified the entire NCCID cohort as a single participant. Unlike the CTs procured from China which are all non-contrast, around 80% of CTs from NCCID are acquired with contrast materials (e.g., iodine). These contrast materials are usually utilized to block X-rays and appeared with higher attenuation on CTs, which could help emphasise tissues such as blood vessels and intestines (in Supplementary Fig. 1 and Table 3). However, in practice, we found that a simple combination of the contrast and the non-contrast CTs did not back the training of a well-generalized model since their intrinsic differences induced in the acquisition procedures[21]. Therefore, to overcome the data heterogeneity between the contrast and non-contrast CTs in the NCCID, we applied an unpaired image-to-image translation method called CycleGAN[22] to transform the contrast CTs into non-contrast variants as augmentations during the local model training. In Supplementary Table 4, we have compared CycleGAN with two other recent image translation methods (CouncilGAN[23] and ACL-GAN[22]). We showed that the model trained on CycleGAN transformed contrast CTs has the best performance (test on the non-contrast CTs). However, this modality transformation is not always helpful, as the performance degenerated when training on the raw plus translated contrast CTs.

We developed a densely connected 3D convolutional neural network (CNN) model based on the massive cohort collection towards delivering precise diagnoses with AI approaches. We term it 3D-DenseNet and report its architectural designs and training optimisations in the Methods and Supplementary Fig. 2. We examined the predictive power of 3D-DenseNet on a four-class pneumonia classification task as well as COVID-19 identification. In the first task, we aimed at distinguishing COVID-19 (Fig. 3a, Supplementary Fig. 3 and Table 5) from healthy cases and two other pneumonia types, namely non-COVID-19 viral and bacterial pneumonia (Fig. 3b). We preferred a four-class taxonomy since further distinguishment of COVID-19 with community-acquired pneumonia (CAP)[24,25] can help deliver more commendatory clinical treatments, where the bacterial and the viral are two primary pathogens of CAP[26] (Fig. 2c). However, given different institutions accompanied by various annotating protocols, it is more feasible for the model to learn to discriminate COVID-19 from all non-COVID-19 cases. Therefore, we base the experimental results on this two-category classification in the main text. We report the four-class experiments based on the Wuhan Tongji Hospital Group's cohort in Supplementary Fig. 3 and Table 5.

For the three UCADI data centres in China (Main Campus, Optical Valley and Sino-French branches of Wuhan Tongji Hospital Group), the locally trained 3D-DenseNet achieved an average test sensitivity/specificity of 0.804/0.708 for identifying COVID-19. As for the collection from Britain (NCCID), with the help of CycleGAN to mitigate the heterogeneity between contrast and non-contrast CTs, the test sensitivity/specificity (on non-contrast CTs) of the local model can be improved from 0.703/0.961 to 0.784/0.961. In Supplementary Table 6 and 7, we further compared 3D-DenseNet with two other 3D CNN baseline models: 3D-ResNet[27] and 3D-Xception[28]. As a result, we demonstrated that 3D-DenseNet had better performance and smaller size, presenting it as highly suitable for federated learning.

To interpret the learned features of the model, we performed gradient-weighted class activation mapping (GradCAM)[29] analysis on the CTs from the test set. We visualised the featured regions that lead to identification decisions. It has been found that the generated heatmaps (Fig. 3c) primarily characterised local lesions that are highly overlapped with the radiologists' annotations, suggesting the model is capable of learning robust radiologic features rather than simply overfitting[30]. This heatmap can help the radiologists localise the lesions quicker for delivering diagnoses in an actual clinical environment. Moreover, localising the lesions will also provide a guide for further CT acquisition and clinical test. A similar idea has been described as "region-of-interest (ROI) detection" in a previous similar study[31].

To examine the cross-domain generalisation ability of the locally trained models, we tested China's locally trained model on Britain's test set and vice versa. We reported the numerical results in Fig. 4. However, due to incompleteness, isolation, and heterogeneity in the various data resources, we found that all the locally trained models exhibited less-than-ideal test performances on other sources. Specifically, the model trained on NCCID non-contrast CTs had a sensitivity/specificity/AUC of 0.313/0.907/0.745 in identifying COVID-19 on the test set of China, which is lower than locally trained ones, and vice versa. Next, we describe how to incorporate federated learning for the cross-continent privacy-preservation collaboration on training a generalised AI diagnostic model, mitigating the domain gaps and data heterogeneity.

- **Enable multination privacy-preserving collaboration with federated learning**

We developed a federated learning framework to facilitate the collaboration nested under UCADI and NCCID, integrating diverse cohorts as part of a global joint effort on developing a precise and robust AI diagnostic tool. In traditional data science approaches[17,31], sensitive and private data from different sources are directly gathered and transported to a central hub where the models are deployed. However, such procedures are infeasible in real clinical practises; hospitals are usually reluctant (and often not permitted) for data disclosure due to privacy concerns and legislation[32]. On the other side, the federated learning technique proposed by Google[33], in contrast, is an architecture where the AI model is distributed to and executed at each host institution without data centralisation. Furthermore, transmitting the model parameters effectively reduced the latency and the cost associated with sending large amounts of data during internet connections. More importantly, the strategy to preserve privacy by design enables medical centres to collaborate on developing models without sharing sensitive clinical data with other institutions. Recently, Swarm Learning[34] is proposed towards the model decentralisation via edge computation. However, we conjecture it is immature for the privacy-preserving machine learning[35] applications based on massive data collection and participants due to the exponential increase in computation.

In UCADI, we have provided: (i) An online diagnostic interface allowing people to query the diagnostic results on identifying COVID-19 by uploading their chest CTs; (ii) A federated learning framework that enables UCADI participants to collaboratively contribute to improving the AI model for COVID-19 identification. Each UCADI participant will send the model weights back to the server via a customised protocol during the collaborative training process every few iterations. To further mitigate the potential for data leaks during such a transmission process, we applied an additive homomorphic encryption method called Learning with Errors (LWE) [36] to encrypt the transmitted model parameters. By so doing, participants will keep within their data and infrastructure, with the central server having no access whatsoever. After receiving the transmitted packages from the UCADI participants, the central server then aggregates the global model without comprehending the model parameters of each participant. The updated global model would then be distributed to all participants, again utilising LWE encryption, enabling the continuation of the model optimisation at the local level. Our framework is designed to be highly flexible, allowing dynamic participation and breakpoint resumption (detailed in Methods).

With this framework, we deployed the same experimental configurations to validate the federated learning concept for developing a generalized CT-based COVID-19 diagnostic model (detailed in Methods). We compared the test sensitivity and specificity of the federated model to the local variations (Fig. 4). We plotted the ROC curves and calculated the corresponding AUC scores, along with 95% confidence intervals (CI) and p-values, to validate the model's performance (Fig. 4). As confirmed by the curves and numbers, the federated model outperformed all the locally trained ones on the same test splits collected from China and the UK. Specifically, for the test performance on the 1,076 CTs of 254 cases in China (all from the three branches of Wuhan Tongji Hospital Group), the federated model achieved a sensitivity/specificity/AUC of 0.973/0.951/0.980, respectively, outperforming the models trained locally at Main Campus, Optical Valley, Sino-French and NCCID. In addition, the federated model achieves a sensitivity/specificity/AUC of 0.730/0.942/0.894 for COVID-19 classification when applied to the test set of the NCCID (from 18 UK hospitals), vastly outperforming all the locally trained models. We based the performance measure on the CT level instead of the patient level, coherent with the prior study[31].

We illustrated that the federated framework is an effective solution to mitigate against the issue that we cannot centralise medical data from hospitals worldwide due to privacy and legal legislation. We further conducted a comparative study on the same task with a panel of expert radiologists. With an average of 9 years' experience, six qualified radiologists from the Department of Radiology, Wuhan Tongji Hospital (Main Campus), were asked to make diagnoses on each CT from China, as one of the four classes. The six experts were first asked to provide diagnoses individually, then to address integrated diagnostic opinions via majority votes (consensus) in a plenary meeting. We presented the radiologists and AI models with the same data partition for a fair comparison. In differentiating COVID-19 from the non-COVID-19 cases, the six radiological experts obtained an average 0.79

in sensitivity (0.88, 0.90, 0.55, 0.80, 0.68, 0.93, respectively), and 0.90 in specificity (0.92, 0.97, 0.89, 0.95, 0.88, 0.79, respectively). In reality, the consideration of a clinical decision is usually made by consensus decision among the experts. Here, we use the majority votes among the six expert radiologists to represent such a decision-making process. We provide the detailed diagnostic decisions of each radiologist in Supplementary Table 5. We found that the majority vote helps reduce the potential bias and risk: the aggregated diagnoses are with the best performance among individual radiologists. In Fig. 4a, we plotted the majority votes in blue markers (sensitivity/specificity: 0.900/0.956) and remarked that the federatively trained 3D-DenseNet had shown comparable performance (sensitivity/specificity: 0.973/0.951) with the expert panel. We have further presented and discussed the models' performance on the hold-out test sets (645 cases from Wuhan Tianyou Hospital and 506 cases from Wuhan Union Hospital) in Supplementary Table 8. We proved that the federatively trained model also performed better on these two hold-out datasets, yet the confidence sometimes is not well calibrated.

During the federated training process, each participant is required to synchronise the model weights with the server every few training epochs using web sockets. Intuitively, more frequent communication should lead to better performance. However, each synchronisation accumulates extra time. To investigate the trade-off between the model performance and the communication cost during the federated training, we conduct parallel experiments with the same settings but different training epochs between the consecutive synchronisations. We report the models' subsequent test performance in Fig. 5a and time usage in Fig. 5b. We observe that, as expected, more frequent communication leads to better performance. Compared with the least frequently communication scenario, to download the model from the beginning and train locally without intermediate communications, synchronizing at every epoch will achieve the best test performance with less than 20% increment in time usage.

## Discussion

COVID-19 is a global pandemic. Over 200 million people have been infected worldwide, with hundreds of thousands hospitalized and mentally affected[37,38], and as of Oct 2021, above four million are reported to have died. There are borders between countries, yet the only barrier is the boundary between humankind and the virus. We urgently demand a global joint effort to confront this illness effectively. In this study, we introduced a multination collaborative AI framework, UCADI, to assist radiologists in streamlining and accelerating CT-based COVID-19 diagnoses. Firstly, we developed a new CNN model that achieved performance comparable to expert radiologists in identifying COVID-19. The predictive diagnoses can be utilised as references while the generated heatmap helps with faster lesion localisation and further CT acquisition. Then, we formed a federated learning framework to enable the global training of a CT-based model for precise and robust diagnosis. With CT data from 22 hospitals, we have herein confirmed the effectiveness of the federated learning approach. We have shared the trained model and open-sourced the federated learning framework. It is worth mentioning that our proposed framework is with continual evolution, is not confined to the diagnosis of COVID-19 but also provides infrastructures for future use. The uncertainty and heterogeneity are the characteristics of clinical work. Because of the limited medical understanding of the vast majority of diseases, including pathogenesis, pathological process, treatment, etc., the medical characteristics of diseases can be studied by the means of AI. Along with this venue, research can be more instructive and convenient in dealing with large (sometimes isolated) samples, especially suitable for transferring knowledge in studying emerging diseases.

However, certain limitations are not well addressed in this study. First is the potential bias in the comparison between experts and models. Due to legal legislation, it is infeasible and impossible to disclose the UK medical data with radiologists and researchers in China or vice versa. Thus, radiologists are from nearby institutions. Though their diagnostic decisions are quite different, it is not unrealistic to conclude that our setting and evaluation process eliminate biases. The Second is engineering efforts. Although we have developed mechanisms such as dynamic participation and breakpoint resumption, the participants still happened to drop from the federated training process for the unstable internet connection. Also, the computation efficiency of the 3D CNN model still has space for improvements (in Supplementary Table 7). There are always engineering advancements that can be incorporated to refine the framework.

## Methods

We first described how we constructed the dataset, then we discussed the details of our implementations for collaboratively training the AI model, we provided further analysis of our methods at the end of this section.

- **CN dataset development (UCADI)**

A total of 5,740 chest CT images that are acquired from the three branches (Main Campus, Optical Valley and Sino-French) of Tongji Hospital Group located in Wuhan, China, using similar acquisition protocols. Three scanners are used to obtain the CTs: GE Medical System/LightSpeed16, GE Medical Systems/Discovery 750 HD and Siemens SOMATOM Definition AS+. The scanning slice thickness is set as 1.25mm and 1mm for the GE and the Siemens scanners, respectively. The reconstruction protocols include a statistical iteration (60%) and sinogram affirmed iteration for the GE and the Siemens devices, respectively. All the Chinese-derived CTs are taken without the intravenous injection of iodine contrast agent (i.e., non-contrast CTs). Regarding the acquisition date, 2,723 CTs of the 432 COVID-19 patients were enrolled, selected, and annotated from January 7, 2020; 3,017 CTs from other three categories were then retrieved from the databases of these three hospitals, with an event horizon going back to 2016.

As detailed in Supplementary Information, the chest CTs were then divided into a training/validation (hereafter: trainval) split of 1,095 cases, and a testing split of 254 cases. The trainval split consists of 342 cases (1,136 CTs) for healthy individuals, 405 cases (2,200 CTs) for those COVID-19 positive, 56 cases (250 CTs) for other viral pneumonia and 292 cases (1,078 CTs) for bacterial pneumonia. For the test split, we considered a balanced distribution over the four classes, consisting of 80 cases (262 CTs) for healthy individuals, 94 cases (523 CTs) for the COVID-19 positive instances, 20 cases (84 CTs) for other viral pneumonia and 60 cases (207 CTs) for bacterial pneumonia. Specifically, the virus types that are regarded as "other viral pneumonia" include respiratory syncytial, Epstein–Barr, cytomegalovirus, influenza A and parainfluenza.

Additionally, we collected independent cohorts including 507 COVID-19 cases from Wuhan Union Hospital and 645 COVID-19 cases from Wuhan Tianyou Hospital. These hold-out test sets were used for testing the generalisation of the locally trained models as well as the federated model. Since the data source only contained COVID-19 cases, we did not utilize it during the training process. We also summarised and reported the demographic information (i.e., gender and age) of the cohort in Supplementary Table 1.

- **UK dataset development (NCCID)**

For the total 2,682 CTs that were acquired from the 18 partner hospitals located in the United Kingdom (See Supplementary Table 3), the acquisition devices and protocols varied from hospital to hospital. There are over 14 types of utilised CT scanners: Siemens Sensation 64; Siemens SOMATOM Drive; Siemens SOMATOM Definition AS/AS+/Edge/Flash; GE Medical Systems Optima CT660; GE Medical Systems Revolution CT/EVO; GE Medical Systems LightSpeed VCT; Canon Medical Systems Aquilion ONE; Philips Ingenuity Core 128 and Toshiba Aquilion ONE/PRIME. Settings such as filter sizes, slice thickness and reconstruction protocols are also quite diverse among these CTs. This might explain the reason why the NCCID locally trained model failed to perform as well as the Chinese locally trained variant (see Fig. 4c). Regarding the material differences, 2,145 out of 2,682 CTs were taken after the injection of an iodine contrast agent (i.e., contrast CTs). As pointed out by previous study[21], contrast and non-contrast CTs have different feature distributions in terms of attenuation and brightness; it is therefore infeasible to simply mix all the CTs together for local or federated training. The reported numbers in Fig. 3 are based on the non-contrast CTs, while in Supplementary Table 3, we used CycleGAN[22] to incorporate both contrast and non-contrast CTs, and shall elaborate upon such settings in the following section.

As detailed in Supplementary Information, CTs from NCCID were first partitioned into two types: contrast and non-contrast. Such division is based on the metadata provided in the CTs as well as validated from the professional radiologists. For the contrast CTs, the trainval produces a split of 421 cases, and a testing split of 243 cases. The trainval split consists of 276 cases (1,097 CTs) for non-COVID-19 and 145 cases (491 CTs) for the COVID-19 positive cases. The test split contains 160 cases (259 CTs) for non-COVID-19 and 83 cases (138 CTs) for the COVID-19 positives. The non-contrast CTs is fewer in quantity compared with the contrast ones. It has 116 cases (394 CTs) for non-COVID-19 and 54 cases (163 CTs) for the COVID-19 positive cases. Moreover, there are 75 cases (103 CTs) for non-COVID-19 and 27 cases (37 CTs) for the COVID-19 positive cases for the test split.

We also noticed that a small subset of the CTs only contained partial lung regions, we removed these insufficient CTs whose number of slices are less than 40. As for our selection criteria in this regard, although the partial lung scans might be infeasible for training segmentation or detection models, we believe that a sufficient number of slices is enough to ensure the model effectively captures the requisite features and thereby help with the precise classification in medical diagnosis.

We reported patient demographical information (i.e., gender and age) of the cohort in Supplementary Table 2. However, the reported demographics is not inclusive since the demographical attributes of non-COVID-19 cases are not recorded. In comparison to the demographical information of the COVID-19 cases acquired from China, COVID-19 cases in the UK were with larger averaged ages and had more male patients. These demographical differences might also explain why the UK locally trained model failed to perform well when applied to the CTs acquired from China.

- **Data pre-processing, model architecture and training setting**

We pre-processed the raw acquired CTs for standardisation as well as to reduce the burden on computing resource. We utilized an adaptive sampling method to select 16 slices from all sequential images of a single CT case using random starting positions and scalable transversal intervals. During the training and validation process, we sampled once for each CT study, while in testing we repeated the sampling five independent times to obtain five different subsets. We then standardised the sampled slices by removing the channel-wise offsets and rescaling the variation to uniform units. During testing, the five independent subsets of each case were fed to the trained CNN classifier to obtain the prediction probabilities of the four classes. We then averaged the predictive probabilities over these five runs to make the final diagnostic prediction for that case. By so doing, we can effectively include impacts from different levels of lung regions as well as to retain scalable computations. To further improve the computing efficiency, we utilised trilinear interpolation to resize each slice from 512 to 128 pixels along each axis and rescaled the lung windows to a range between -1200 and 600 Hounsfield units before feeding into the network model.

We named our developed model 3D-DenseNet (Supplementary Fig. 2). It was developed based on DenseNet[39], a densely connected convolutional network model that performed remarkably well in classifying 2D images. To incorporate such design with the 3D CT representations, we adaptively customized the model architecture into fourteen 3D-convolution layers distributed in six dense blocks and two transmit blocks (insets of Supplementary Fig. 2). Each dense block consists of two 3D convolution layers and an inter-residual connection, whereas the transmit blocks are composed of a 3D convolution layer and an average pooling layer. We placed a 3D DropBlock[40] instead of simple dropout[41] before and after the six dense blocks, which proved to be more effective in regularising the training of convolution neural networks. We set the momentum of batch normalisation[42] to be 0.9, and the negative slope of LeakyReLU activation as 0.2.

During training, the 3D-DenseNet took the pre-processed CT slice sequences as the input, then output a prediction score over the four possible outcomes (pneumonia types). Due to the data imbalance, we defined the loss function as the weighted cross entropy between predicted probabilities and the true categorical labels. The weights were set as 0.2, 0.2, 0.4, 0.2 for healthy, COVID-19, other viral pneumonia, and bacterial pneumonia cases, respectively. We utilised SGD optimiser with a momentum of 0.9 to update parameters of the network via backpropagation. We trained the networks using a batch size of 16. At the first five training epochs, we linearly increased the learning rate to the initial set value of 0.01 from zero. This learning rate warm-up heuristic proved to be helpful, since using a large learning rate at the very beginning of the training may result in numerical instability[43]. We then used cosine annealing[44] to decrease the learning rate to zero over the remaining 95 epochs (100 epochs in total).

During both local and federated training processes, we utilized a five-fold cross-validation on trainval split, and then selected the best model and reported their test performance (in Fig. 4 and Supplementary Fig. 2).

- **Federated learning and privacy preservation**

At the central server, we adapted the FedAvg[33] algorithm to aggregate the updated model parameters from all clients (i.e., UCADI participants), that is, to combine the weights with respect to clients' dataset sizes and the number of local training epochs between consecutive communications. To ensure secure transmissions between the server and the clients, we used an encryption method called "Learning with Errors" (LWE)[36] to further protect all the transmitted information (i.e., model parameters and metadata). LWE is an additively homomorphic variant of the public key encryption scheme, therefore the participant information cannot even leak to the server, which is to say, that the server has no access to the explicit weights of the model. Compared with other encryption methods, such as differential privacy (DP)[45], Moving Horizon Estimations (MHE)[46] and Model Predictive Control (MPC)[47], LWE differentiates itself by essentially enabling the clients to achieve identical performance with the variants trained without decryption. However, the LWE method would add additional costs to the federated learning framework in terms of the extra encryption/decryption process and the increased size of the encrypted parameters during transmission. The typical time usage of a single encryption-decryption round is 2.7s (average over 100 trials under a test environment consisting of a single CPU (Intel Xeon E5-2630 v3 @ 2.40GHz) and the encrypted model size arises from 2.8MB to 62 MB, which increases the transmission time from 3.1s to 68.9s, in a typical international bandwidth environment[48] of 900KB/s (Fig. 5).

- **Comparing with professional radiologists**

We further conducted a comparative study on this four-type classification between the CNN model and expert radiologists. We asked six qualified radiologists (average of 9 years of clinical experience, range from 4 to 18 years) from the Tongji Hospital Group, to make the diagnoses based on the CTs. We provided the radiologists with the CTs and their labels from the China-derived trainval split. We then asked them to diagnose each CT from the test split into one of the four classes. We reported the performance of each single radiologist and the majority votes on the COVID-19 vs non-COVID-19 CTs in Fig. 4 (detailed comparisons are presented in Supplementary

Table 5 and 9). If there are multiple majority votes for different classes, the radiologist panel will make further discussions until reaching a consensus.

- **Augmented contrast/non-contrast CTs with CycleGAN**

Following similar procedures as previous work[21], we first extracted and converted the slices from contrast and non-contrast CTs of NCCID into JPEG format images with a resolution of 512px by 512px. The trainval and test splits of the contrast CTs contain 932 images (23 cases) and 139 images (22 cases), respectively. For the non-contrast CTs, there are 1,233 images (26 cases) and 166 images (26 cases) for the trainval and test splits, respectively. For the architecture of the CycleGAN, we use ResNet[49] backbone as the feature encoder and set the remaining parts in concordance with the original literature[21]. For the training settings of CycleGAN, we used a batch size of 12 for the total number of 200 epochs. We used the same settings on the trade-off coefficients in the adversarial loss. We started with a learning rate of 2e-4, kept it constant for the first 100 epochs, then decayed it to zero linearly over the next 100 epochs.

To evaluate the effectiveness of utilising CycleGAN for augmentation, we first trained the 3D DenseNet on trainval set of: (i) only non-contrast; (ii) non-contrast and CycleGAN synthesized non-contrast; (iii) only contrast and (iv) contrast and CycleGAN synthesized contrast CTs. In Supplementary Table 3, we reported the test performance of these trained models on the non-contrast and contrast CTs respectively. We observed that augmenting the non-contrast CTs with CycleGAN would result in a better identification ability of the model while this was not held when converting the non-contrast ones into contrast.

- **Data availability**

The clinical data collected from the 23 hospitals that utilised in this study remains under their custody. Part of the data are available via applications from qualified teams. Please refer to the NCCID website (https://www.nhsx.nhs.uk/covid-19-response/data-and-covid-19/national-covid-19-chest-imaging-database-nccid/) for more details.

- **Code availability**

The online application to join UCADI is provided at http://www.covid-ct-ai.team. Codes are publicly available at: https://github.com/HUST-EIC-AI-LAB/UCADI, ref[50], which is released under a Creative Commons Attribution-NonCommercial 3.0 Unported License (CC BY-NC 3.0).

**Acknowledgements**

This study was supported by these grants: HUST COVID-19 Rapid Response Call (No.2020kfyXGYJ021, No. 2020kfyXGYJ031, No. 2020kfyXGYJ093, No. 2020kfyXGYJ094), the National Natural Science Foundation of China (61703171 and 81771801), National Cancer Institute, National Institutes of Health, U01CA242879, and Thammasat University Research fund under the NRCT, Contract No. 25/2561, for the project of "Digital platform for sustainable digital economy development", based on the RUN Digital Cluster collaboration scheme. H.W acknowledges support from Cambridge Trust, Kathy Xu Fellowship, Centre for Advanced Photonics and Electronics and the Cambridge Philosophical Society; M.R acknowledge support from AstraZeneca, Intel and the DRAGON consortium; A.W. acknowledges support from a Turing AI Fellowship under grant EP/V025379/1, The Alan Turing Institute, and the Leverhulme Trust via CFI. C.B.S. acknowledges support from DRAGON, Intel, the Philip Leverhulme Prize, the Royal Society Wolfson Fellowship, the EPSRC grants EP/S026045/1 and EP/T003553/1, EP/N014588/1, EP/T017961/1, the Wellcome Innovator Award RG98755, the Leverhulme Trust project Unveiling the invisible, the European Union Horizon 2020 research and innovation programme under the Marie Skodowska-Curie grant agreement No. 777826 NoMADS, the Cantab Capital Institute for the Mathematics of Information and the Alan Turing Institute. The funders did not participate in the research or review any details of this study.

**Author Contributions Statement**

T.X., and X.B. conceived the work. H.W., J.G., Z.F. F.Y., and K.M. contributed to the design and development of the models, software, and the experiments. T.X., X.B., H.W., J.G., Z.F., K.M., J.L., M.R., and C.S. interpreted, analysed, and presented the experimental results. T.X., H.W., J.G., Z.F., K.M., X.B., Y.X., W.L., A.W., J.L., M.R., and C.S. contributed to drafting and revising the manuscript. All the authors contributed to the data acquisition and resource allocation.

**Competing Interests Statement**

The authors declare no competing interests.

**Ethics Approval**

The UK data used in this study is under approval by Control of patient information (COPI) notice issued by The Secretary of State for Health and Social Care. The CN data usage is approved by the Ethics Committee Tongji Hospital, Tongji Medical College of Huazhong University of Science and Technology.

**Tables**

We don't have tables in the main text.

**Figure Legends/Captions**

Fig. 1 | Conceptual architecture of UCADI. The participants first download and train the 3D CNN models based on the data of local cohorts. The trained model parameters are then encrypted and transmitted back to the server. Finally, the server produces the federated model via aggregating the contributions from each participant while without explicit access to the parameters.

Fig. 2 | Deployment and workflow of UCADI participants. a, Data. Construct a local dataset based on the high-quality, well-annotated and anonymised CTs. b, Flow. The backbone of the 3D DenseNet model mainly consists of six 3D dense blocks (in green), two 3D transmit blocks (in white), and an output layer (in grey). CTs of each case are converted into a (16,128,128) tensor after adaptive sampling, decentralisation and trilinear interpolation, then feed into the 3D CNN model for pneumonia classification. c, Process. During training, the model outputs are used to calculate the weighted cross entropy to update the network parameters. While testing, five independent predictions of each case are incorporated to report the predictive diagnostic results.

Fig. 3 | Overview of CTs. a, Radiological features correlated with COVID-19 pneumonia cases: ground glass opacity, interlobular septal thickening and consolidation (from left to right). b, Other non-COVID-19 cases, incl. healthy, other viral and bacterial pneumonia. c, Localised class-discriminative regions generated by GradCAM (in the heatmap) and annotated by the professional radiologists (circled in red), for COVID-19 cases.

Fig. 4 | COVID-19 pneumonia identification performance of 3D CNN models trained on four different data resources (Main Campus, Optical Valley, Sino-French and NCCID) individually and federatively. a, Receiver Operating Characteristic (ROC) curves when the models are tested on the data from China, in comparison with

six professional radiologists, b, ROC curves of the CNN models tested on the data from the UK, c, Numeric results of the test sensitivity, specificity and area under the curve (AUC, with 95% confidence intervals and p-values)

Fig. 5 | Trade-off on the performance and communication cost in federated training. a, Relationships between transmission expense and model generalisation, b, Estimated time spent at different communication/synchronisation intervals. The statistics is measured based on a joint FL training of two clients. Each client has 200 CTs and 100 CTs for training and testing, respectively. The client's software infrastructure is a single-core of GPU (NVIDIA GTX 1080Ti) and a CPU (Xeon(R) CPU E5-2660 v4 @ 2.00GHz). The bandwidth for transmission is around 7.2Mb/s (900KB/s), which is the average broadband speed.

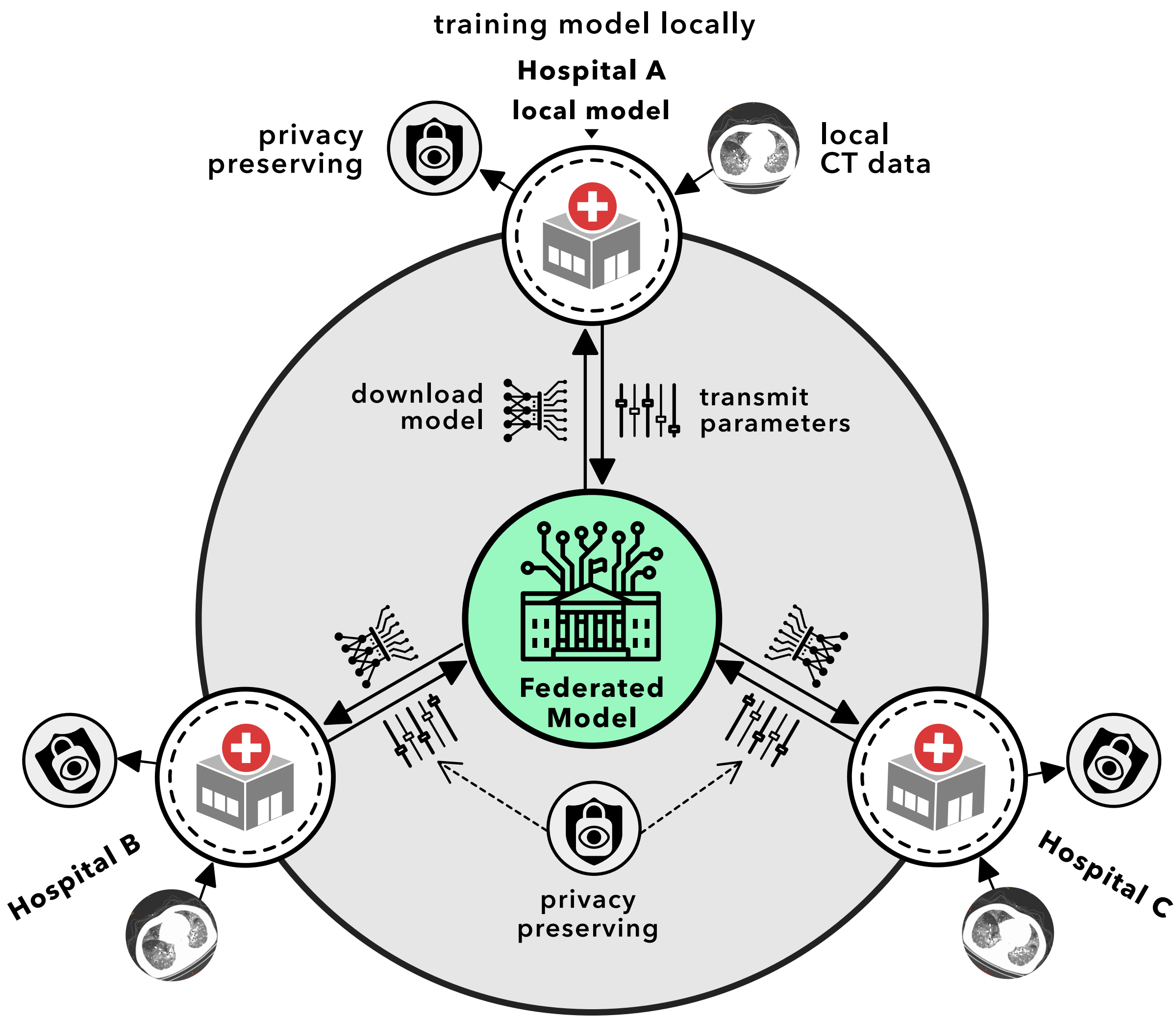

**Fig. 1 | Conceptual architecture of UCADI.** The participants first download and train the 3D CNN models based on the data of local cohorts. The trained model parameters are then encrypted and transmitted back to the server. Finally, the server produces the federated model via aggregating the contributions from each participant while without explicit access to the parameters.

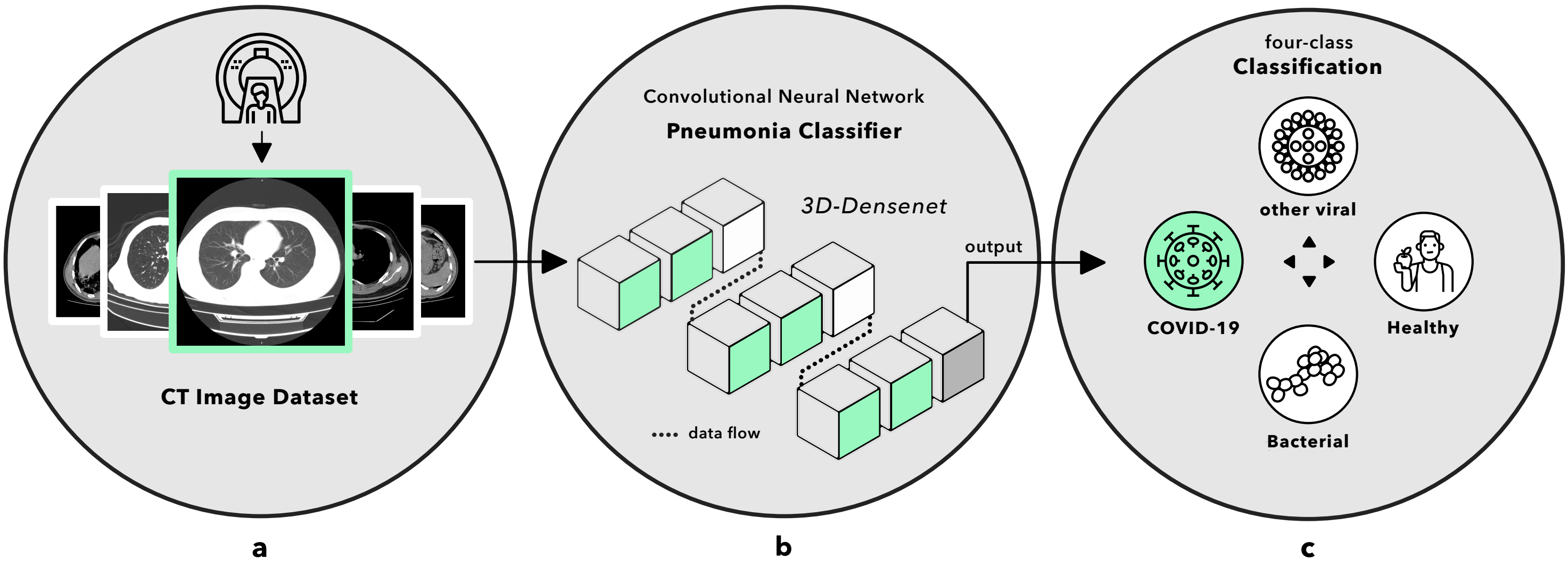

**Fig. 2 | Deployment and workflow of UCADI participants. a**, Data. Construct a local dataset based on the high-quality, well-annotated and anonymised CTs. **b**, Flow. The backbone of the 3D DenseNet model mainly consists of six 3D dense blocks (in green), two 3D transmit blocks (in white), and an output layer (in grey). CTs of each case are converted into a (16,128,128) tensor after adaptive sampling, decentralisation and trilinear interpolation, then feed into the 3D CNN model for pneumonia classification. **c**, Process. During training, the model outputs are used to calculate the weighted cross entropy to update the network parameters. While testing, five independent predictions of each case are incorporated to report the predictive diagnostic results.

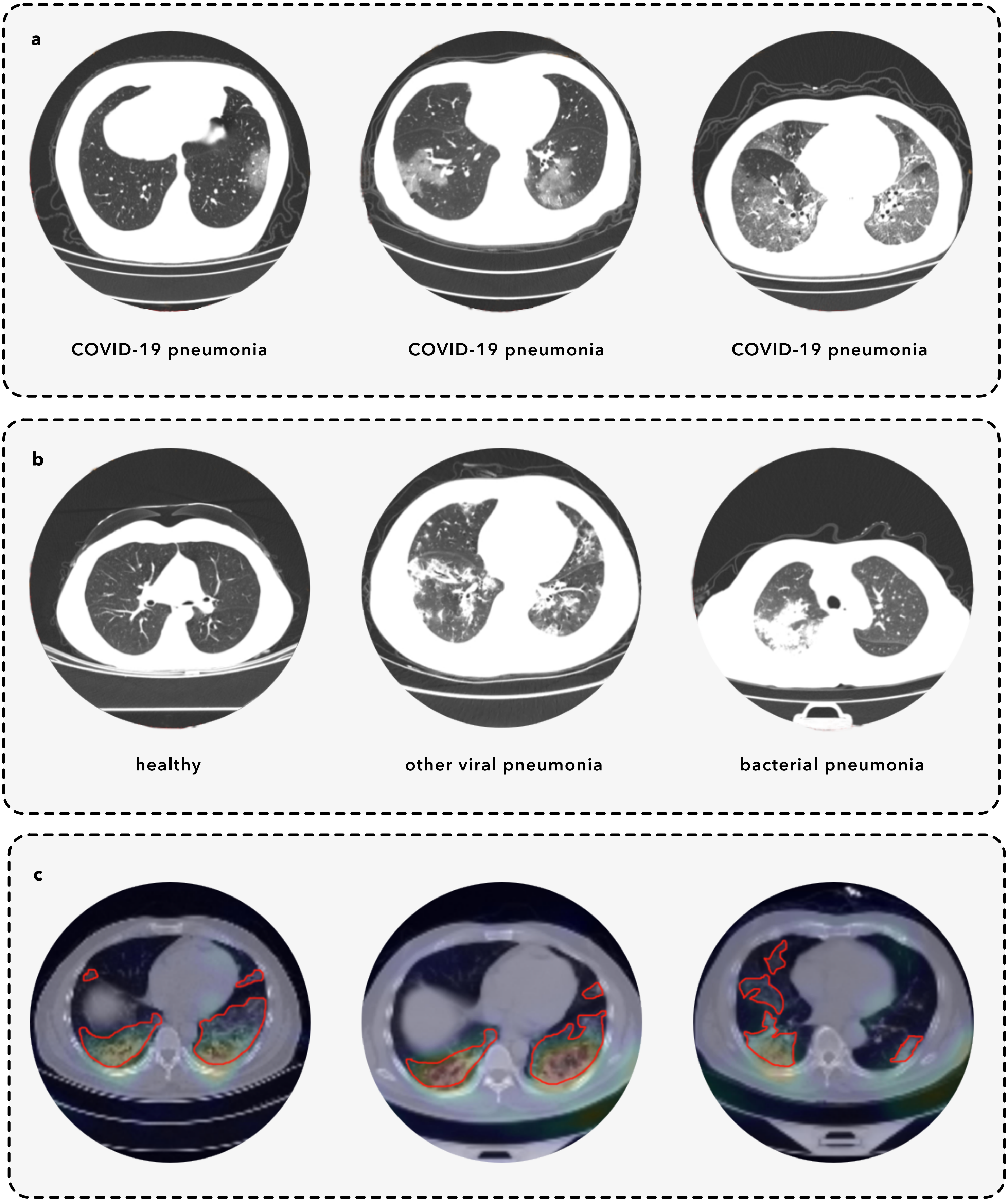

**Fig. 3 | Overview of CTs. a,** Radiological features correlated with COVID-19 pneumonia cases: ground glass opacity, interlobular septal thickening and consolidation (from left to right). **b,** Other non-COVID-19 cases, incl. healthy, other viral and bacterial pneumonia. **c,** Localised class-discriminative regions generated by GradCAM (in the heatmap) and annotated by the professional radiologists (circled in red), for COVID-19 cases.

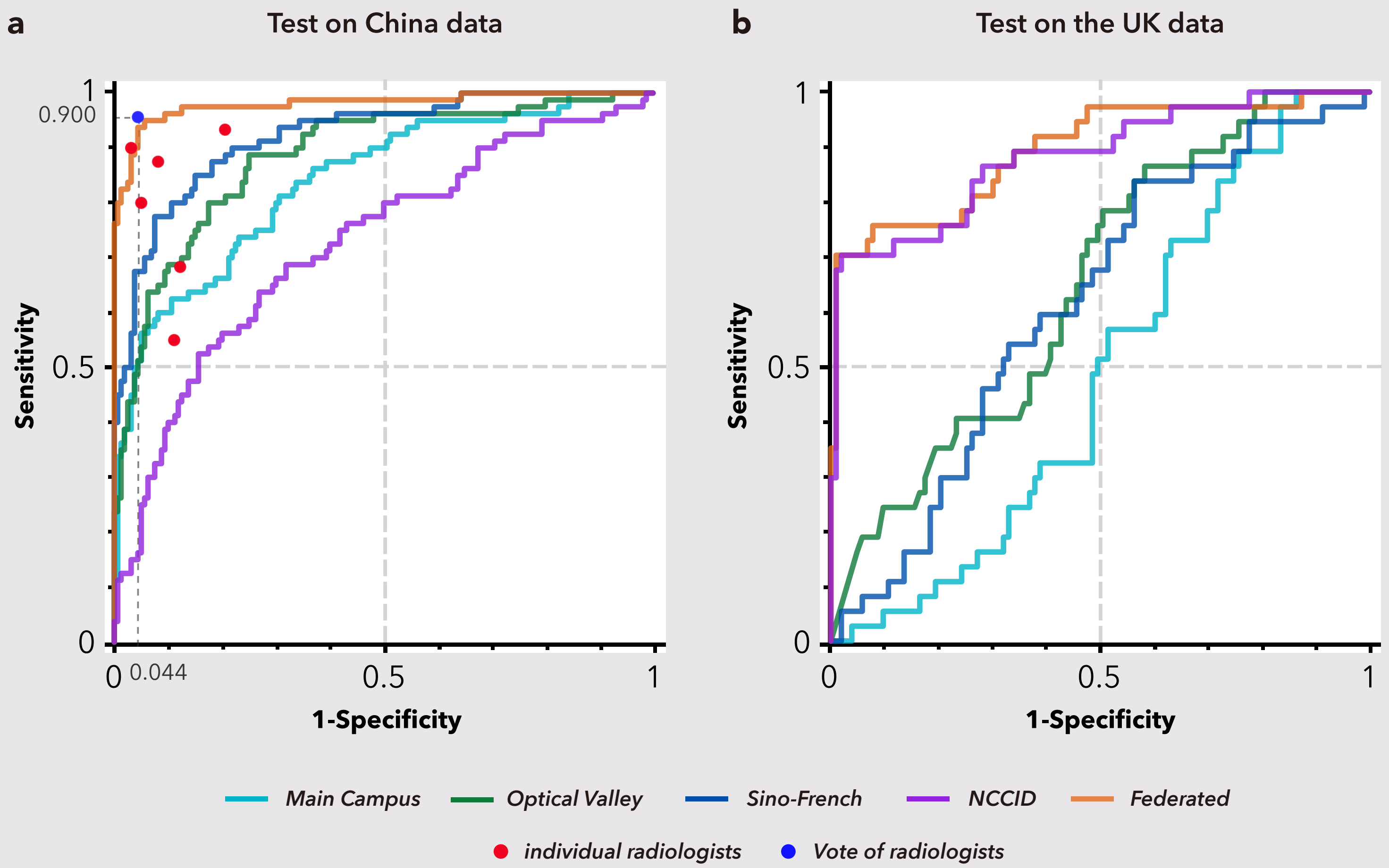

**c**

| | | Main Campus | Optical Valley | Sino-French | NCCID | Federated |
|---|---|---|---|---|---|---|
| China | Sensitivity | 0.538 | 0.973 | 0.900 | 0.313 | 0.973 |
| | Specificity | 0.926 | 0.444 | 0.759 | 0.907 | 0.951 |
| | AUC | 0.840 | 0.884 | 0.922 | 0.745 | 0.980 |
| | 95% CI | [0.788 0.898] | [0.826 0.921] | [0.874 0.947] | [0.657 0.800] | [0.958 0.996] |
| | p-value | 0.022 | 0.019 | 0.016 | 0.028 | 0.009 |
| UK | Sensitivity | 0.054 | 0.541 | 1.000 | 0.703 | 0.730 |
| | Specificity | 0.835 | 0.626 | 0.160 | 0.961 | 0.942 |
| | AUC | 0.487 | 0.647 | 0.613 | 0.882 | 0.894 |
| | 95% CI | [0.438 0.522] | [0.594 0.698] | [0.582 0.631] | [0.822 0.952] | [0.876 0.951] |
| | p-value | 0.032 | 0.021 | 0.005 | 0.027 | 0.014 |

**Fig. 4 | COVID-19 pneumonia identification performance of 3D CNN models trained on four different data resources (Main Campus, Optical Valley, Sino-French and NCCID) individually and federatively. a,** Receiver Operating Characteristic (ROC) curves when the models are tested on the data from China, in comparison with six professional radiologists, **b,** ROC curves of the CNN models tested on the data from the UK, **c,** Numeric results of the test sensitivity, specificity and area under the curve (AUC, with 95% confidence intervals and p-values)

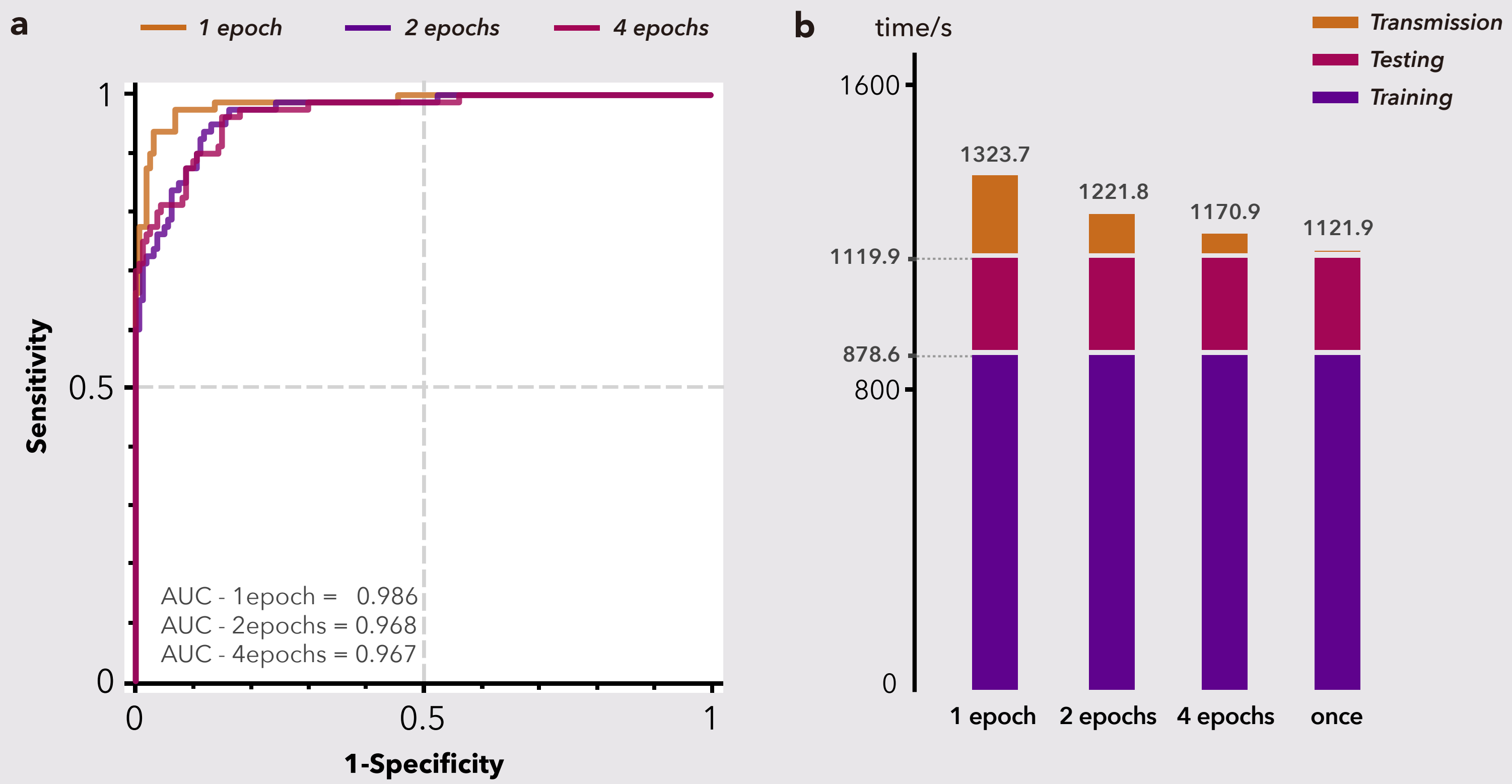

**Fig. 5 | Trade-off on the performance and communication cost in federated training. a,** Relationships between transmission expense and model generalisation, **b,** Estimated time spent at different communication/synchronisation intervals. The statistics is measured based on a joint FL training of two clients. Each client has 200 CTs and 100 CTs for training and testing, respectively. The client's software infrastructure is a single-core of GPU (NVIDIA GTX 1080Ti) and a CPU (Xeon(R) CPU E5-2660 v4 @ 2.00GHz). The bandwidth for transmission is around 7.2Mb/s (900KB/s), which is the average broadband speed.

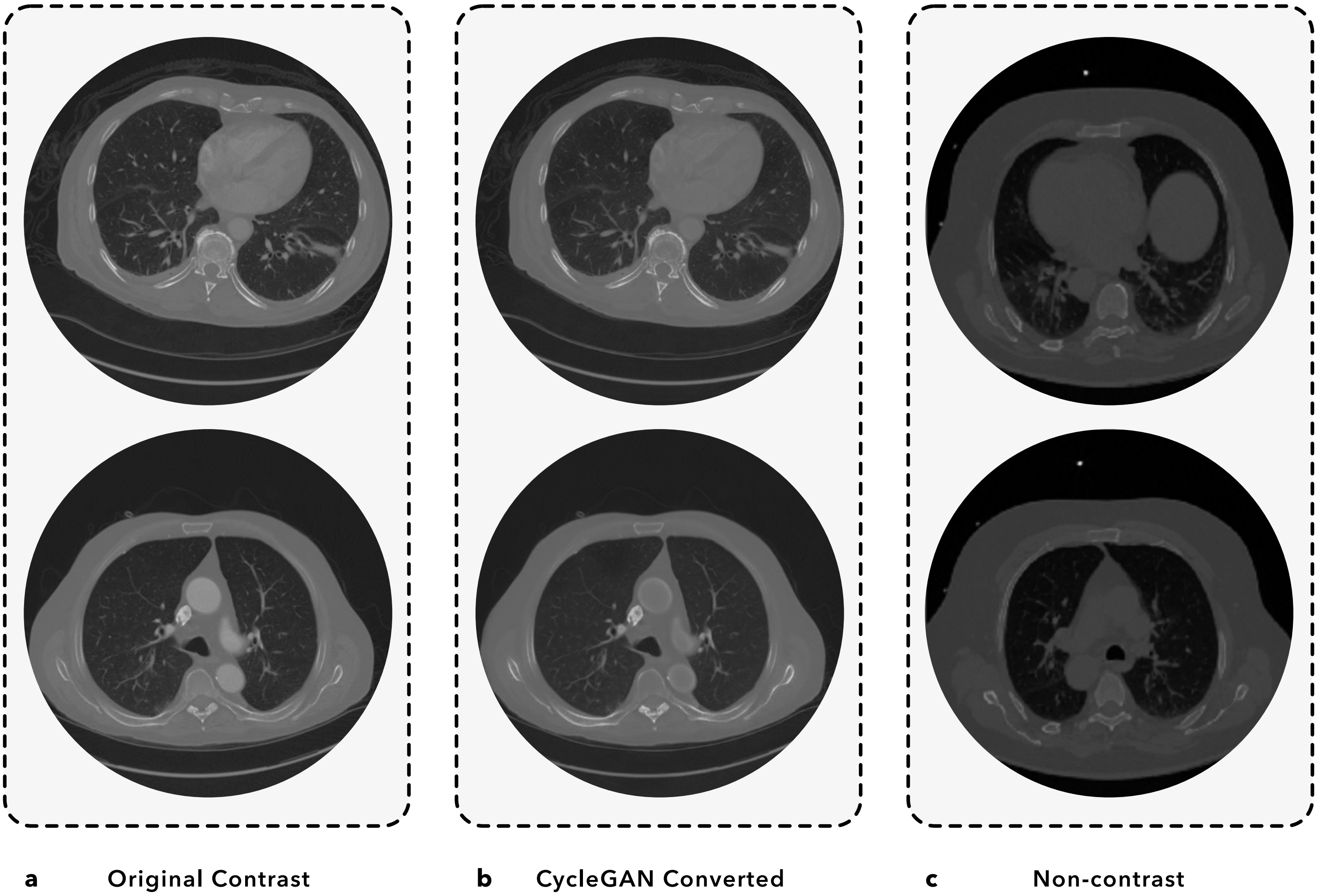

**Supplementary Fig. 1 | Conversion of the contrast CTs with CycleGAN. a,** Contrast CTs (angiography), **b,** Correspondent converted CTs using CycleGAN to reduce the contrastive regions (pointed out by red arrows), **c,** Non-contrast CTs.

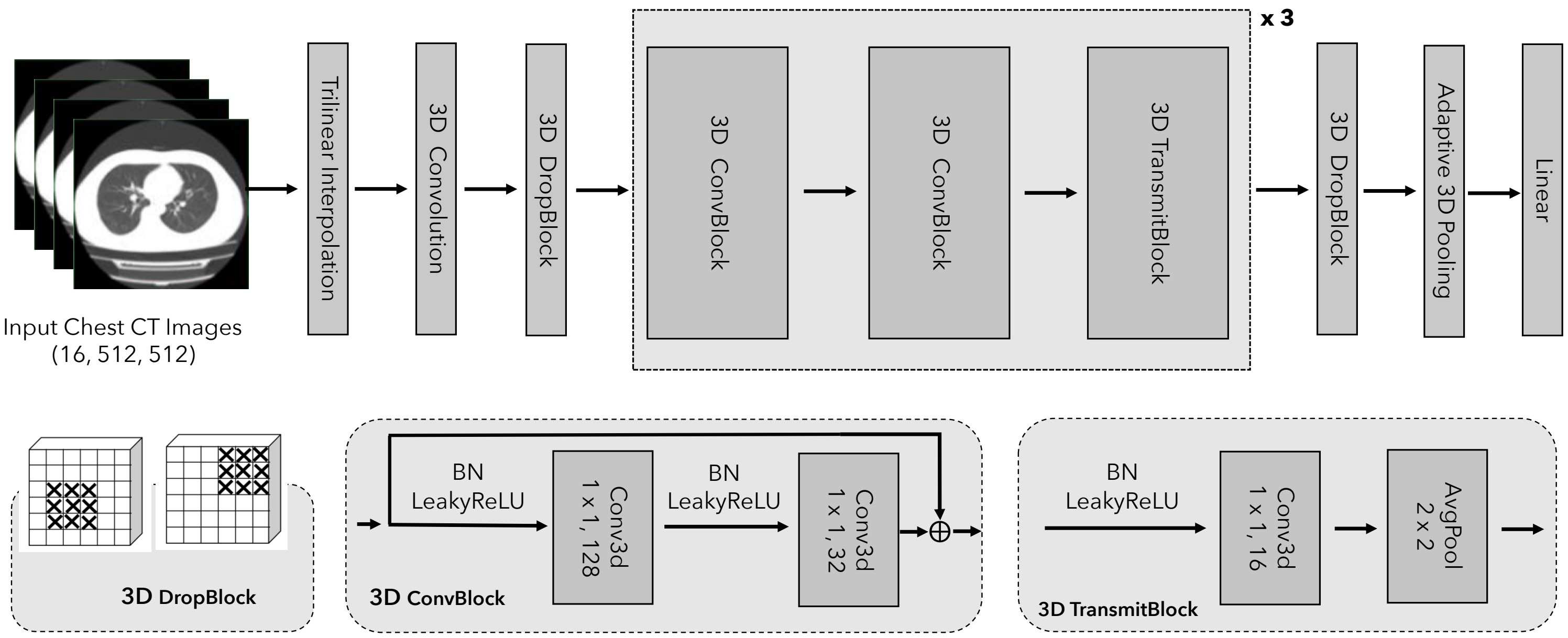

**Supplementary Fig. 2 | Architecture of 3D DenseNet.**

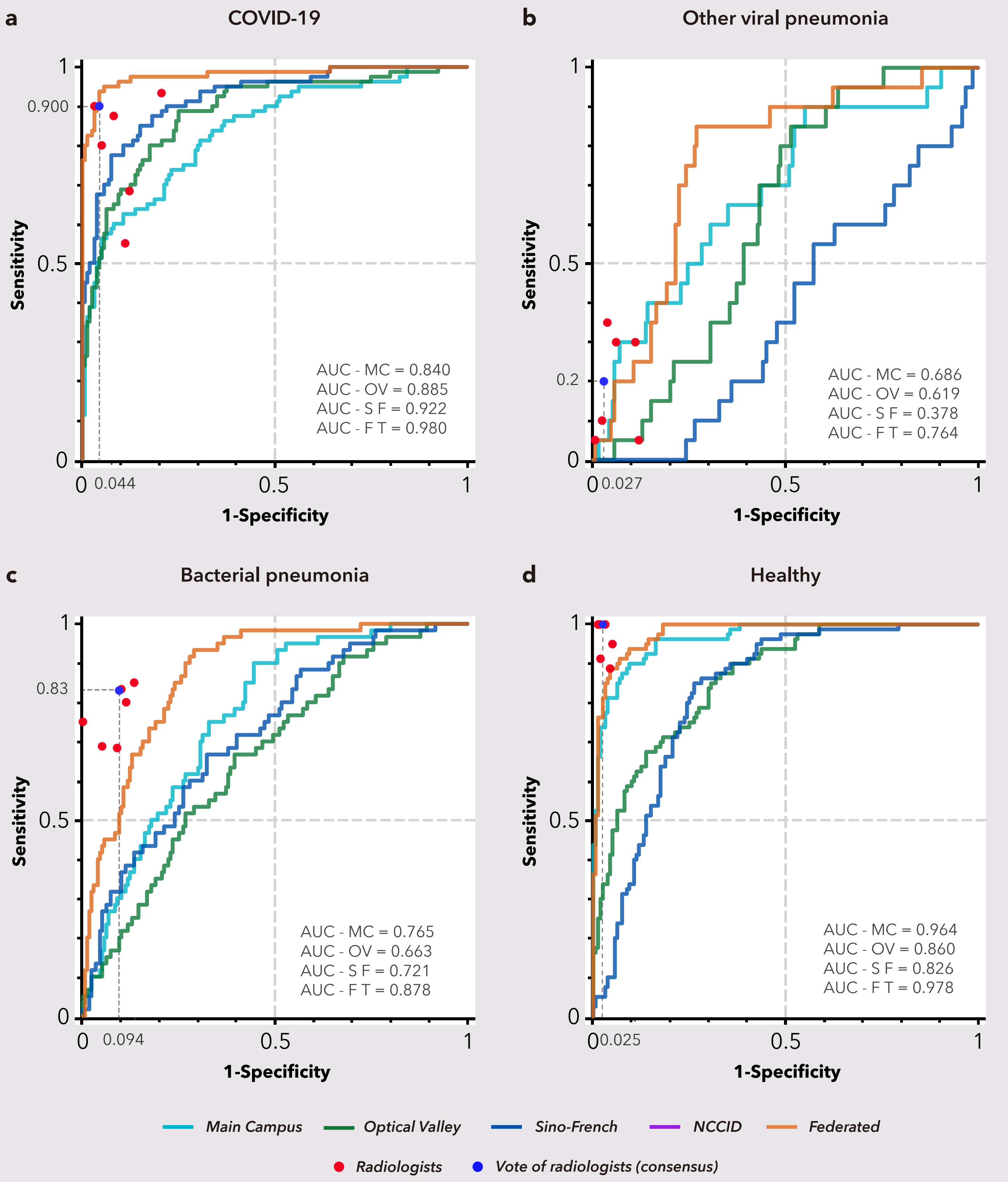

**Supplementary Fig. 3 | Detailed model performance on the four-class (a-d) prediction from the China data.**

| Train split | | | | | | | | | | |
|---|---|---|---|---|---|---|---|---|---|---|
| **Resource** | **Pneumonia cohort** | **Gender** | | **Age** | | | | | **Patient amount** | **CT amount** |
| | | Female | Male | 0-20 | 20-40 | 40-60 | 60-80 | 80+ | | |
| Main Campus | N/A(healthy) | 115 | 109 | 18 | 144 | 59 | 3 | 0 | 224 | 727 |
| | COVID-19 | 66 | 69 | 5 | 39 | 45 | 43 | 3 | 135 | 922 |
| | Other viral | 26 | 30 | 0 | 8 | 29 | 19 | 0 | 56 | 250 |
| | Bacterial | 118 | 136 | 1 | 47 | 104 | 94 | 8 | 254 | 934 |
| Optical Valley | N/A(healthy) | 32 | 43 | 8 | 57 | 10 | 0 | 0 | 75 | 278 |
| | COVID-19 | 54 | 58 | 1 | 16 | 30 | 57 | 8 | 112 | 425 |
| | Other viral | 0 | 0 | 0 | 0 | 0 | 0 | 0 | 0 | 0 |
| | Bacterial | 1 | 12 | 0 | 1 | 3 | 7 | 2 | 13 | 47 |
| Sino-French | N/A(healthy) | 22 | 21 | 1 | 31 | 10 | 1 | 0 | 43 | 131 |
| | COVID-19 | 85 | 73 | 2 | 30 | 53 | 66 | 7 | 158 | 853 |
| | Other viral | 0 | 0 | 0 | 0 | 0 | 0 | 0 | 0 | 0 |
| | Bacterial | 10 | 15 | 0 | 8 | 12 | 5 | 0 | 25 | 97 |
| | | | | | | | | total | **1095** | **4664** |

| Test split | | | | | | | | | | |
|---|---|---|---|---|---|---|---|---|---|---|
| **Resource** | **Pneumonia cohort** | **Gender** | | **Age** | | | | | **Patient amount** | **CT amount** |
| | | Female | Male | 0-20 | 20-40 | 40-60 | 60-80 | 80+ | | |
| Main Campus | N/A(healthy) | 30 | 28 | 3 | 36 | 19 | 0 | 0 | 58 | 191 |
| | COVID-19 | 21 | 13 | 1 | 5 | 15 | 11 | 2 | 34 | 191 |
| | Other viral | 8 | 11 | 0 | 4 | 9 | 6 | 0 | 19 | 72 |
| | Bacterial | 23 | 27 | 0 | 9 | 20 | 21 | 0 | 50 | 170 |
| Optical Valley | N/A(healthy) | 6 | 6 | 0 | 9 | 3 | 0 | 0 | 12 | 44 |
| | COVID-19 | 13 | 10 | 0 | 6 | 5 | 11 | 1 | 23 | 88 |
| | Other viral | 0 | 0 | 0 | 0 | 0 | 0 | 0 | 0 | 0 |
| | Bacterial | 1 | 1 | 0 | 0 | 1 | 1 | 0 | 2 | 8 |
| Sino-French | N/A(healthy) | 5 | 5 | 0 | 9 | 1 | 0 | 0 | 10 | 27 |
| | COVID-19 | 15 | 22 | 0 | 5 | 12 | 19 | 1 | 37 | 244 |
| | Other viral | 1 | 0 | 0 | 0 | 1 | 0 | 0 | 1 | 12 |
| | Bacterial | 2 | 6 | 1 | 1 | 3 | 3 | 0 | 8 | 29 |
| Tianyou | COVID-19 | - | - | - | - | - | - | - | 645 | 645 |
| Union | COVID-19 | - | - | - | - | - | - | - | 506 | 506 |
| | | | | | | | | total | **1405** | **2227** |

**Supplementary Table 1 | Demographical and clinical information of the CN data.**

| Training and validation split | | | | | | | | | | |
|---|---|---|---|---|---|---|---|---|---|---|
| **CT Info** | **Pneumonia cohort** | **Gender** | | **Age** | | | | | **Patient amount** | **CT amount** |
| | | Female | Male | 0-20 | 20-40 | 40-60 | 60-80 | 80+ | | |
| Contrast | N/A(healthy) | 46 | 67 | 0 | 1 | 11 | 62 | 43 | 276 | 1097 |
| | COVID-19 | 56 | 88 | 0 | 4 | 35 | 68 | 37 | 145 | 491 |
| Non-Contrast | N/A(healthy) | 21 | 24 | 0 | 1 | 5 | 29 | 10 | 116 | 394 |
| | COVID-19 | 21 | 31 | 0 | 0 | 7 | 28 | 17 | 54 | 163 |
| | | | | | | | | total | **591** | **2145** |

| Test split | | | | | | | | | | |
|---|---|---|---|---|---|---|---|---|---|---|
| **Resource** | **Pneumonia cohort** | **Gender** | | **Age** | | | | | **Patient amount** | **CT amount** |
| | | Female | Male | 0-20 | 20-40 | 40-60 | 60-80 | 80+ | | |
| Contrast | N/A(healthy) | 29 | 42 | 0 | 0 | 4 | 40 | 27 | 160 | 259 |
| | COVID-19 | 32 | 51 | 0 | 2 | 20 | 39 | 22 | 83 | 138 |
| Non-Contrast | N/A(healthy) | 11 | 18 | 0 | 0 | 4 | 14 | 9 | 75 | 103 |
| | COVID-19 | 11 | 16 | 0 | 1 | 4 | 19 | 5 | 27 | 37 |
| | | | | | | | | total | **345** | **537** |

**Hospitals Include**: Royal United Hospitals Bath NHS Foundation Trust, Brighton and Sussex University Hospitals NHS Trust, London North West University Healthcare NHS Trust, George Eliot Hospital NHS Trust, Cwm Taf Morgannwg University Health Board, Hampshire Hospitals NHS Foundation Trust, Betsi Cadwaladr University Health Board, Ashford and St Peter's Hospitals, Royal Cornwall Hospitals NHS Trust, Sheffield Children's NHS Foundation Trust, Liverpool Heart and Chest Hospital NHS Foundation Trust, Norfolk and Norwich University Hospitals NHS Foundation Trust, Royal Surrey NHS Foundation Trust, Sandwell and West Birmingham NHS Trust, West Suffolk NHS Foundation Trust, Somerset NHS Foundation Trust, Cambridge University Hospitals NHS Foundation Trust, Imperial College Healthcare NHS Trust

**Supplementary Table 2 | Demographical and clinical information of the UK data.** Part of the cohort demographics is not recorded by some NCCID partnership hospitals.

| Test Split | | Training and Validation Split | | | |
|---|---|---|---|---|---|
| | | Non-contrast | | Contrast | |
| | | Real | Real + Synthetic | Real | Real + Synthetic |
| Non-contrast | Sensitivity | 0.703 | 0.784 | 0.324 | |
| | Specificity | 0.961 | 0.961 | 0.864 | |
| | AUC | 0.882 | 0.921 | 0.655 | |
| Contrast | Sensitivity | 0.269 | | 0.914 | 0.810 |
| | Specificity | 0.897 | | 0.916 | 0.933 |
| | AUC | 0.667 | | 0.979 | 0.931 |

**Supplementary Table 3 | COVID-19 pneumonia identification performance of CNN models trained on contrast and non-contrast split of NCCID dataset (UK).** "Real + Synthetic" means the training and validation images include the original split (non-contrast/contrast) as well as the synthesized ones from their couterpart (contrast/non-contrast) via CycleGAN. We conduct no modification on the test set.

| Method | CouncilGAN | ACL-GAN | CycleGAN (ours) |
|---|---|---|---|
| Performance (Sensitivity / Specificity) | 0.458 / 0.692 | 0.524 / 0.279 | 0.703 / 0.784 |

**Supplementary Table 4 | Comparison on the unpaired image translation methods.**

| Radiologist Expert (Name Abbr.) | | ZL | LYM | YZL | COX | HLM | GC | Majority Vote |
|---|---|---|---|---|---|---|---|---|
| Healthy | sensitivity | 1 | 1 | 0.91 | 0.89 | 0.95 | 1 | 1 |
| | specificity | 0.99 | 0.98 | 0.98 | 0.96 | 0.95 | 0.97 | 0.98 |
| COVID-19 | sensitivity | 0.88 | 0.9 | 0.8 | 0.55 | 0.68 | 0.93 | 0.9 |
| | specificity | 0.92 | 0.97 | 0.95 | 0.89 | 0.88 | 0.79 | 0.96 |
| Other viral pneumonia | sensitivity | 0.1 | 0.35 | 0.3 | 0.05 | 0.3 | 0.05 | 0.2 |
| | specificity | 0.97 | 0.96 | 0.94 | 0.88 | 0.89 | 1.0 | 0.97 |
| Bacterial pneumonia | sensitivity | 0.83 | 0.68 | 0.85 | 0.8 | 0.69 | 0.75 | 0.83 |
| | specificity | 0.9 | 0.91 | 0.87 | 0.89 | 0.95 | 1 | 0.91 |
| COVID vs non-COVID | sensitivity | 0.88 | 0.90 | 0.55 | 0.80 | 0.68 | 0.93 | 0.90 |
| | specificity | 0.92 | 0.97 | 0.89 | 0.95 | 0.88 | 0.79 | 0.96 |

**Supplementary Table 5 | Detailed diagnoses from radiologist.**

| Model | 3D-ResNet50 | 3D-ResNet101 | 3D-Xception | 3D-DenseNet (ours) |
|---|---|---|---|---|
| Performance (Sensitivity / Specificity) | 0.888 / 0.500 | 0.894 / 0.512 | 0.831 / 0.525 | **0.926** / **0.538** |
| Model size (# Params) | 46.16 M | 85.21 M | 22.43 M | **0.73 M** |

**Supplementary Table 6 | Comparison on the 3D CNN models.**

| Model | 3D-ResNet50 | 3D-ResNet101 | 3D-Xception | 3D-DenseNet (ours) |
|---|---|---|---|---|
| FLOPs | 20.32G | 30.31G | 35.19G | 143.50G |

**Supplementary Table 7 | Computation cost of 3D CNN models.**

| Hold-out Data | Cohorts (Positive/Negative) | Performance (Sensitivity) | | |
|---|---|---|---|---|
| | | Main Campus | NCCID | FL |
| Wuhan Union Hospital | 506 / 0 | 0.44 | 0.53 | 0.66 |
| Wuhan Tianyou Hospital | 645 / 0 | 0.42 | 0 | 0.99 |

**Supplementary Table 8 | Model generalisation on the hold-out dataset.**

| | Taxonomy (Pneumonia type) | Main Campus | Optical Valley | Sino-French | NCCID | Federated |
|---|---|---|---|---|---|---|
| Four class | healthy - 0 | [18, 11, 2, 49] | [27, 39, 0, 14] | [15, 22, 0, 43] | [68, 12, 0, 0] | [76, 1, 3, 0] |
| | covid - 1 | [0, 45, 0, 35] | [1, 77, 0, 2] | [2, 71, 0, 7] | [29, 51, 0, 0] | [1, 78, 1, 0] |
| | other viral - 2 | [0, 1, 0, 19] | [1, 11, 0, 8] | [1, 5, 0, 14] | [13, 7, 0, 0] | [0, 4, 5, 11] |
| | bacterial - 3 | [0, 20, 0, 58] | [5, 40, 0, 15] | [6, 14, 0, 40] | [41, 19, 0, 0] | [14, 5, 12, 29] |
| Two class | non-covid - 0 | [146, 14] | [70, 90] | [119, 41] | [122, 38] | [150, 10] |
| | covid - 1 | [35, 45] | [3, 77] | [9, 71] | [29, 51] | [2, 78] |

**Supplementary Table 9 | Confusion matrices of locally/federatively trained models.**